\pdfoutput=1
\documentclass[11pt,letterpaper]{article}

\usepackage[english]{babel}
\usepackage[utf8]{inputenc}
\usepackage[T1]{fontenc}
\usepackage{lmodern}
\usepackage[letterpaper,textwidth=6.0in,textheight=8.6in,centering]{geometry}
\usepackage{amsmath,amsthm,amssymb,mathtools}
\usepackage{booktabs}
\usepackage{microtype}
\usepackage{enumitem}
\usepackage{array}
\usepackage{tikz}
\usetikzlibrary{arrows.meta,positioning,calc}
\usepackage{algorithm}
\usepackage{algpseudocode}
\usepackage[numbers,sort&compress]{natbib}
\usepackage[colorlinks=true,linkcolor=blue!50!black,citecolor=blue!50!black,urlcolor=blue!50!black]{hyperref}

\hypersetup{
  pdftitle={Activation Sensitivity as a Unifying Principle for Post-Training Quantization},
  pdfauthor={Bruce Changlong Xu},
  pdfsubject={Machine Learning (cs.LG); Computation and Language (cs.CL)}
}

\theoremstyle{plain}
\newtheorem{theorem}{Theorem}[section]

\newtheorem{proposition}[theorem]{Proposition}

\theoremstyle{definition}

\theoremstyle{remark}

\newcommand{\Tr}{\operatorname{Tr}}
\newcommand{\diag}{\operatorname{diag}}
\newcommand{\vecop}{\operatorname{vec}}
\newcommand{\quant}{\operatorname{quant}}
\newcommand{\E}{\mathbb{E}}
\newcommand{\R}{\mathbb{R}}

\title{\textbf{Activation Sensitivity as a Unifying Principle\\[0.15em]
for Post-Training Quantization}}

\author{
  Bruce Changlong Xu \\[0.35em]
  {\normalsize Department of Computer Science, Stanford University} \\[0.15em]
  {\small \texttt{brucechanglongxu@cs.stanford.edu}}
}

\date{}

\begin{document}
\maketitle

\begin{abstract}
\noindent Post-training quantization (PTQ) methods for large language models compress pretrained weights to low bit-widths without retraining by optimizing layer-local proxy objectives. Activation-aware scaling methods such as AWQ protect input channels with large activation norms, whereas second-order error-compensation methods such as GPTQ redistribute rounding error using the inverse of the unweighted input covariance matrix. We show that both paradigms---along with diagonal Fisher quantization, block reconstruction, and Hadamard rotation methods---are special cases of minimizing a single loss-derived quadratic form governed by \emph{gradient-weighted activation sensitivity}. From a Gauss--Newton expansion of the task loss at a linear layer $Y = X W^\top$, we prove that under isotropic output curvature the expected excess loss from weight perturbation $\Delta W$ reduces to $\frac{1}{2 d_{\mathrm{out}}}\Tr(\Delta W \, H_D \, \Delta W^\top)$, where $H_D = \frac{1}{n} X^\top D X$ weights each calibration token's outer product by its squared output-gradient norm $D_{tt} = \|\nabla_{y_t}\mathcal{L}\|_2^2$. Within this framework, AWQ corresponds to a diagonal approximation of $H_D$ with uniform token weights ($D = I_n$) optimized via channel scaling, while GPTQ corresponds to Cholesky error compensation on $H_D$ under $D = I_n$. We derive closed-form optimal scaling exponents for diagonal sensitivity protection, formulate sensitivity-weighted Cholesky quantization (SAQ-GPTQ) on $H_D$, and prove bounds on token-gradient heterogeneity bias, cross-layer error amplification across residual blocks, and finite-sample calibration concentration. Controlled numerical simulations confirm that under token-gradient heterogeneity, replacing $X^\top X$ with $X^\top D X$ preserves $>99\%$ top-channel saliency recall and reduces INT3/INT4 excess loss by roughly $3\times$ over unweighted GPTQ.

\vspace{0.65em}
\noindent\textbf{Keywords:} Post-training quantization, large language models, activation sensitivity, Gauss--Newton Hessian, Kronecker-factored curvature, AWQ, GPTQ, Optimal Brain Surgeon.
\end{abstract}

\section{Introduction}
\label{sec:intro}

Autoregressive inference in large language models (LLMs) is memory-bandwidth bound during token generation, making weight-only post-training quantization (PTQ) to 3 or 4 bits per parameter a standard deployment technique \citep{dettmers2022llmint8,frantar2023gptq,lin2024awq}. Unlike quantization-aware training \citep{jacob2018integeronly,choi2018pact,esser2020lsq,zhou2016dorefa}, which fine-tunes discretized weights via end-to-end backpropagation over large training corpora, PTQ must discretize billions of parameters in a single pass using only a few hundred calibration sequences. Because evaluating the global network loss after every weight rounding decision is computationally intractable, practical PTQ algorithms decompose compression into independent layer-wise or block-wise subproblems.

Across recent LLM quantization literature, two seemingly distinct layer-local principles have achieved strong empirical performance:
\begin{enumerate}[label=(\arabic*),topsep=2pt,itemsep=2pt]
  \item \textbf{Activation-magnitude scaling and outlier isolation (AWQ, SmoothQuant, SpQR):}
  Observing that transformer activations above 6B parameters concentrate large magnitudes in a small subset of input feature channels \citep{dettmers2022llmint8}, methods such as AWQ \citep{lin2024awq} rank input channels by their activation norm and apply per-channel weight scaling $W_{:,j} \mapsto s_j W_{:,j}$ prior to group-wise round-to-nearest (RTN) quantization. Related methods migrate scale factors between activations and weights \citep{xiao2023smoothquant} or isolate outlier channels and weights via channel splitting or sparse-quantized representations \citep{zhao2019ocs,dettmers2022llmint8,dettmers2024spqr}.
  \item \textbf{Second-order covariance error compensation (AdaRound, OBQ, GPTQ, QuIP):}
  Rooted in Optimal Brain Damage \citep{lecun1990optimalbraindamage} and Optimal Brain Surgeon \citep{hassibi1993optimalbrainsurgeon}, methods such as AdaRound \citep{nagel2020adaround}, OBC/OBQ \citep{frantar2022obc}, GPTQ \citep{frantar2023gptq}, ZeroQuant \citep{yao2022zeroquant}, and QuIP \citep{chee2023quip,tseng2024quipsharp} formulate layer quantization as minimizing unweighted output reconstruction error $\frac{1}{n}\|X W^\top - X \widehat{W}^\top\|_F^2 = \Tr\bigl(\Delta W (\frac{1}{n}X^\top X) \Delta W^\top\bigr)$. In OBC and GPTQ, as each column $j$ of $W$ is quantized, the resulting rounding error is compensated across the remaining unquantized columns $k > j$ using the Cholesky factorization of $(X^\top X + \lambda I)^{-1}$.
\end{enumerate}

Despite their shared goal of preserving end-to-end model quality, activation-scaling and second-order reconstruction methods are typically presented as competing paradigms. Activation magnitude measures how strongly an input channel is excited, but ignores both inter-channel correlations and whether downstream layers are sensitive to perturbations in the resulting output directions. Conversely, unweighted layer reconstruction $\frac{1}{n}\|X\Delta W^\top\|_F^2$ accounts for input feature correlations via $X^\top X$, but treats every calibration token $t \in \{1,\dots,n\}$ and every output dimension $i \in \{1,\dots,d_{\mathrm{out}}\}$ as equally important to the task loss. Meanwhile, diagonal Fisher methods such as SqueezeLLM \citep{kim2024squeezellm} discard input-channel off-diagonal correlations in favor of per-weight squared gradients, whereas Hessian-spectrum and hardware-aware mixed-precision methods \citep{wang2019haq,dong2019hawq,dong2020hawqv2,shen2020qbert} summarize entire layers or blocks by their top Hessian eigenvalue or trace for coarse bit-width allocation.

In this paper, we resolve this fragmentation by deriving a unified mathematical framework for post-training quantization centered on \emph{gradient-weighted activation sensitivity}. Specifically:
\begin{itemize}[topsep=2pt,itemsep=2pt]
  \item \textbf{Kronecker-factored Gauss--Newton derivation of activation sensitivity (Section~\ref{sec:formulation}):}
  We analyze the effect of a layer weight perturbation $\Delta W \in \R^{d_{\mathrm{out}} \times d_{\mathrm{in}}}$ on the end-to-end task loss $\mathcal{L}(W)$. Distinguishing carefully between token-coupled first-order perturbations and the second-order Gauss--Newton / empirical Fisher curvature \citep{martens2015kfac}, we prove (Theorem~\ref{thm:sensitivity-matrix}) that under isotropic per-token output curvature, the expected excess loss reduces to the quadratic form $\frac{1}{2}\Tr(\Delta W \, H_D \, \Delta W^\top)$, where
  \[
    H_D = \frac{1}{n} X^\top D X \in \R^{d_{\mathrm{in}} \times d_{\mathrm{in}}}, \qquad
    D = \diag\bigl(\|G_{1,:}\|_2^2, \dots, \|G_{n,:}\|_2^2\bigr),
  \]
  and $G_{t,:} = \nabla_{y_t}\mathcal{L}$ is the backpropagated output gradient on token $t$. The diagonal entries $\alpha_j = (H_D)_{jj} = \frac{1}{n}\sum_{t=1}^n \|G_{t,:}\|_2^2 X_{t,j}^2$ define the \emph{activation sensitivity} of input channel $j$.
  \item \textbf{Unification of existing PTQ paradigms and optimal scaling/compensation laws (Section~\ref{sec:unification}):}
  We show (Proposition~\ref{prop:unification} and Table~\ref{tab:taxonomy}) that AWQ, GPTQ, SqueezeLLM, BRECQ \citep{li2021brecq}, and rotation-based incoherence methods (QuIP\# \citep{tseng2024quipsharp}, QuaRot \citep{ashkboos2024quarot}) correspond to specific structural approximations of $H_D$. Using this formulation, we prove a closed-form optimal scaling exponent $\gamma^* = \frac{1}{p+2}$ ($\gamma^* \in [1/4, 1/3]$ for $\ell_p$ group dynamic-range envelopes with $p \in [1, 2]$) balancing quantization step-size inflation against high-sensitivity channel protection (Theorem~\ref{thm:optimal-scaling}), and we derive sensitivity-weighted Cholesky error compensation (Theorem~\ref{thm:saq-gptq} and Algorithm~\ref{alg:saq-gptq}).
  \item \textbf{Theoretical limits of static layer-local quantization (Section~\ref{sec:limitations}):}
  We prove three results quantifying when unweighted layer-local proxies fail: (i) an exact covariance identity for the bias $\alpha_j - \frac{1}{n}\|X_{:,j}\|_2^2 \bar{d}$ under token-gradient heterogeneity (Proposition~\ref{prop:heterogeneity-bias}), (ii) a Lipschitz recurrence for cross-layer quantization drift across $L$-block residual transformers (Theorem~\ref{thm:cross-layer}), and (iii) a finite-sample concentration bound for calibrating $H_D$ under heavy-tailed activations (Proposition~\ref{prop:calibration-sample}).
  \item \textbf{Reproducible numerical verification (Section~\ref{sec:experiments}):}
  We validate our theoretical predictions in controlled numerical experiments across varying token-gradient heterogeneity $\sigma_g$ and bit-widths (INT4 and INT3). While unweighted GPTQ minimizes unweighted layer reconstruction error, replacing $X^\top X$ with the gradient-weighted sensitivity matrix $H_D = \frac{1}{n}X^\top D X$ preserves $>99.4\%$ top-channel saliency recall at high heterogeneity (where AWQ recall drops to $38.9\%$) and reduces end-to-end excess loss by $3.2\times$ at INT4 and $2.8\times$ at INT3 over unweighted GPTQ.
\end{itemize}

\section{Mathematical Formulation of Activation Sensitivity}
\label{sec:formulation}

\subsection{Setup and Notation}

Consider a pretrained neural network with parameters $\Theta$ evaluated on a calibration dataset of $n$ tokens (or sequence positions). Focus on a single linear layer with weight matrix $W \in \R^{d_{\mathrm{out}} \times d_{\mathrm{in}}}$, which maps stacked input activations $X \in \R^{n \times d_{\mathrm{in}}}$ to pre-activation outputs
\begin{equation}
  Y = X W^\top \in \R^{n \times d_{\mathrm{out}}}, \qquad
  y_t = W x_t \in \R^{d_{\mathrm{out}}} \quad (t = 1, \dots, n),
  \label{eq:linear-layer}
\end{equation}
where $x_t = X_{t,:}^\top \in \R^{d_{\mathrm{in}}}$ and $y_t = Y_{t,:}^\top \in \R^{d_{\mathrm{out}}}$ denote the input activation and output vector for calibration token $t \in \{1, \dots, n\}$, and $X_{:,j} \in \R^n$ denotes the vector of activations across all $n$ tokens in input channel $j \in \{1, \dots, d_{\mathrm{in}}\}$.

Post-training quantization replaces $W$ with a discretized weight matrix $\widehat{W} = Q(W) \in \R^{d_{\mathrm{out}} \times d_{\mathrm{in}}}$, introducing a deterministic weight perturbation
\[
  \Delta W := \widehat{W} - W \in \R^{d_{\mathrm{out}} \times d_{\mathrm{in}}}.
\]
Perturbing a single input channel $j \in \{1, \dots, d_{\mathrm{in}}\}$ modifies only the $j$-th column $\Delta W_{:,j} \in \R^{d_{\mathrm{out}}}$, which induces the rank-one layer output perturbation
\begin{equation}
  \Delta Y_j = X_{:,j} (\Delta W_{:,j})^\top \in \R^{n \times d_{\mathrm{out}}}, \qquad
  \Delta y_{t,j} = X_{t,j} \Delta W_{:,j} \in \R^{d_{\mathrm{out}}}.
  \label{eq:channel-output-pert}
\end{equation}
Taking the Frobenius norm of~\eqref{eq:channel-output-pert} over the $n$ calibration tokens yields
\begin{equation}
  \frac{1}{n} \|\Delta Y_j\|_F^2
  = \frac{1}{n} \Tr\bigl(X_{:,j} (\Delta W_{:,j})^\top \Delta W_{:,j} X_{:,j}^\top\bigr)
  = \|\Delta W_{:,j}\|_2^2 \cdot \left(\frac{1}{n}\|X_{:,j}\|_2^2\right).
  \label{eq:act-mag-derivation}
\end{equation}
Equation~\eqref{eq:act-mag-derivation} is the exact algebraic motivation behind activation-aware quantization \citep{lin2024awq}: if one measures the effect of quantizing channel $j$ solely by the unweighted Euclidean norm of the layer output change $\Delta Y_j$, then channel importance is proportional to the mean squared activation magnitude $\frac{1}{n}\|X_{:,j}\|_2^2$. However,~\eqref{eq:act-mag-derivation} ignores how the rest of the network transforms $\Delta Y_j$ into changes in the task loss.

\subsection{First-Order Token Perturbations vs.\ Second-Order Loss Curvature}

Let $\mathcal{L}(W) = \frac{1}{n} \sum_{t=1}^n \ell_t(y_t)$ denote the average task loss (such as autoregressive cross-entropy) over the $n$ calibration tokens, and let
\[
  G \in \R^{n \times d_{\mathrm{out}}}, \qquad
  g_t := G_{t,:}^\top = \nabla_{y_t} \ell_t(y_t) \in \R^{d_{\mathrm{out}}}
\]
denote the matrix of backpropagated output gradients with respect to $Y$. By the chain rule, the gradient of $\mathcal{L}$ with respect to $W$ is $\nabla_W \mathcal{L}(W) = \frac{1}{n} G^\top X \in \R^{d_{\mathrm{out}} \times d_{\mathrm{in}}}$, whose $j$-th column is $\frac{1}{n} G^\top X_{:,j}$.

For a single token $t \in \{1, \dots, n\}$, a channel perturbation $\Delta W_{:,j}$ shifts the token loss $\ell_t(y_t)$ to first order by
\begin{equation}
  \Delta \ell_{t,j}^{(1)}
  := \langle g_t, \Delta y_{t,j} \rangle
  = X_{t,j} \langle g_t, \Delta W_{:,j} \rangle.
  \label{eq:token-first-order}
\end{equation}
If we sum~\eqref{eq:token-first-order} coherently over the batch before squaring, we obtain $\Delta \mathcal{L}_j^{(1)} = \frac{1}{n}\langle G^\top X_{:,j}, \Delta W_{:,j}\rangle$, whose squared coefficient $\|G^\top X_{:,j}\|_2^2 = X_{:,j}^\top G G^\top X_{:,j}$ mixes cross-token gradient inner products $\langle g_t, g_s\rangle$ for $t \neq s$. At a converged pretrained checkpoint, however, the batch-averaged gradient $\frac{1}{n}G^\top X$ is near zero because positive and negative token gradients cancel across the dataset, even when individual token losses $\ell_t(y_t)$ experience large perturbations!

To capture the true degradation in token-level predictions without cancellation across unrelated tokens, we must either evaluate the mean squared per-token first-order variation $\frac{1}{n}\sum_{t=1}^n (\Delta \ell_{t,j}^{(1)})^2$---which equals the empirical Fisher quadratic form---or expand $\mathcal{L}(W + \Delta W)$ to second order:
\begin{equation}
  \mathcal{L}(W + \Delta W) - \mathcal{L}(W)
  = \langle \nabla_W \mathcal{L}(W), \Delta W \rangle
  + \frac{1}{2} \vecop(\Delta W)^\top \nabla_W^2 \mathcal{L}(W) \vecop(\Delta W)
  + O(\|\Delta W\|_F^3),
  \label{eq:second-order-taylor}
\end{equation}
where $\vecop(\Delta W) \in \R^{d_{\mathrm{out}} d_{\mathrm{in}}}$ stacks the rows of $\Delta W$. Using Kronecker-factored curvature \citep{martens2015kfac}, we now establish the exact relationship between~\eqref{eq:token-first-order},~\eqref{eq:second-order-taylor}, and the gradient-weighted activation sensitivity matrix $H_D$.

\begin{theorem}[Kronecker-Factored Sensitivity and the Matrix $H_D$]
\label{thm:sensitivity-matrix}
Let $W \in \R^{d_{\mathrm{out}} \times d_{\mathrm{in}}}$ be a linear layer $y_t = W x_t$ evaluated on $n$ calibration tokens with inputs $X \in \R^{n \times d_{\mathrm{in}}}$ and output loss gradients $g_t = \nabla_{y_t}\ell_t(y_t) \in \R^{d_{\mathrm{out}}}$. Define the diagonal token-gradient weight matrix
\begin{equation}
  D := \diag\bigl(\|g_1\|_2^2, \|g_2\|_2^2, \dots, \|g_n\|_2^2\bigr) \in \R^{n \times n},
  \label{eq:D-def}
\end{equation}
and the \emph{gradient-weighted activation sensitivity matrix}
\begin{equation}
  H_D := \frac{1}{n} X^\top D X = \frac{1}{n} \sum_{t=1}^n \|g_t\|_2^2 \, x_t x_t^\top \in \R^{d_{\mathrm{in}} \times d_{\mathrm{in}}}.
  \label{eq:HD-def}
\end{equation}
Then:
\begin{enumerate}[label=(\alph*),topsep=2pt,itemsep=2pt]
  \item \textbf{Empirical Fisher Kronecker structure:} Under the empirical Fisher / Gauss--Newton approximation $\nabla_{y_t}^2 \ell_t(y_t) \approx g_t g_t^\top$, the layer Hessian $\mathcal{H}_W \in \R^{(d_{\mathrm{out}} d_{\mathrm{in}}) \times (d_{\mathrm{out}} d_{\mathrm{in}})}$ is given by
  \begin{equation}
    \mathcal{H}_W = \frac{1}{n} \sum_{t=1}^n (g_t g_t^\top) \otimes (x_t x_t^\top),
    \label{eq:kronecker-hessian}
  \end{equation}
  and the quadratic loss perturbation equals the mean squared token-level variation:
  \begin{equation}
    \frac{1}{2} \vecop(\Delta W)^\top \mathcal{H}_W \vecop(\Delta W)
    = \frac{1}{2n} \sum_{t=1}^n \bigl(\langle g_t, \Delta W x_t \rangle\bigr)^2.
    \label{eq:fisher-quad}
  \end{equation}
  \item \textbf{Reduction to $H_D$ under isotropic output gradient directions:} Suppose that conditional on the token gradient norm $\|g_t\|_2$ and input $x_t$, the unit gradient direction $u_t = g_t / \|g_t\|_2$ is isotropic ($\E[u_t u_t^\top \mid \|g_t\|_2, x_t] = \frac{1}{d_{\mathrm{out}}} I_{d_{\mathrm{out}}}$), or equivalently that $\Delta W$ is a zero-mean perturbation with $\E[\Delta W^\top g_t g_t^\top \Delta W \mid x_t, \|g_t\|_2] = \frac{\|g_t\|_2^2}{d_{\mathrm{out}}} \E[\Delta W^\top \Delta W]$. Then~\eqref{eq:fisher-quad} simplifies exactly to
  \begin{equation}
    \E[\Delta \mathcal{L}(\Delta W)]
    = \frac{1}{2 d_{\mathrm{out}}} \Tr\bigl(\Delta W \, H_D \, \Delta W^\top\bigr).
    \label{eq:trace-HD}
  \end{equation}
  \item \textbf{Channel-wise activation sensitivity:} Specializing~\eqref{eq:trace-HD} to a single input channel perturbation $\Delta W_{:,j}$ yields
  \begin{equation}
    \E[\Delta \mathcal{L}(\Delta W_{:,j})]
    = \frac{1}{2 d_{\mathrm{out}}} \|\Delta W_{:,j}\|_2^2 \cdot \alpha_j,
    \qquad\text{where}\quad
    \alpha_j := (H_D)_{jj} = \frac{1}{n} \sum_{t=1}^n \|g_t\|_2^2 \, X_{t,j}^2.
    \label{eq:alpha-j-def}
  \end{equation}
\end{enumerate}
\end{theorem}
\begin{proof}
For part~(a), since $y_t = W x_t = (I_{d_{\mathrm{out}}} \otimes x_t^\top) \vecop(W)$, the Jacobian of $y_t$ with respect to $\vecop(W)$ is $J_t = I_{d_{\mathrm{out}}} \otimes x_t^\top$. Substituting the empirical Fisher approximation $\nabla_{y_t}^2 \ell_t(y_t) \approx g_t g_t^\top$ into the Gauss--Newton Hessian $\mathcal{H}_W = \frac{1}{n}\sum_{t=1}^n J_t^\top \nabla_{y_t}^2 \ell_t(y_t) J_t$ and applying the mixed-product property $(A \otimes B)(C \otimes D) = (AC) \otimes (BD)$ gives
\[
  J_t^\top (g_t g_t^\top) J_t
  = (I_{d_{\mathrm{out}}} \otimes x_t) (g_t g_t^\top \otimes 1) (I_{d_{\mathrm{out}}} \otimes x_t^\top)
  = (g_t g_t^\top) \otimes (x_t x_t^\top),
\]
establishing~\eqref{eq:kronecker-hessian}. Using the identity $\vecop(A)^\top (B \otimes C) \vecop(A) = \Tr(A C A^\top B^\top)$ with $A = \Delta W$, $B = g_t g_t^\top$, and $C = x_t x_t^\top$ yields
\[
  \vecop(\Delta W)^\top \bigl((g_t g_t^\top) \otimes (x_t x_t^\top)\bigr) \vecop(\Delta W)
  = \Tr\bigl(\Delta W x_t x_t^\top \Delta W^\top g_t g_t^\top\bigr)
  = \bigl(g_t^\top \Delta W x_t\bigr)^2,
\]
which averages over $t = 1, \dots, n$ to give~\eqref{eq:fisher-quad}.

For part~(b), writing $g_t = \|g_t\|_2 u_t$ and taking the conditional expectation of $g_t g_t^\top$ given $(\|g_t\|_2, x_t)$ under $\E[u_t u_t^\top \mid \|g_t\|_2, x_t] = \frac{1}{d_{\mathrm{out}}} I_{d_{\mathrm{out}}}$ replaces $\mathcal{H}_W$ by
\[
  \overline{\mathcal{H}}_W
  = \frac{1}{n} \sum_{t=1}^n \left(\frac{\|g_t\|_2^2}{d_{\mathrm{out}}} I_{d_{\mathrm{out}}}\right) \otimes (x_t x_t^\top)
  = \frac{1}{d_{\mathrm{out}}} I_{d_{\mathrm{out}}} \otimes \Biggl(\frac{1}{n}\sum_{t=1}^n \|g_t\|_2^2 x_t x_t^\top\Biggr)
  = \frac{1}{d_{\mathrm{out}}} I_{d_{\mathrm{out}}} \otimes H_D.
\]
Applying $\vecop(\Delta W)^\top (I_{d_{\mathrm{out}}} \otimes H_D) \vecop(\Delta W) = \Tr(\Delta W \, H_D \, \Delta W^\top)$ yields~\eqref{eq:trace-HD}.

Finally, for part~(c), when $\Delta W$ is supported only on column $j$, $\Tr(\Delta W \, H_D \, \Delta W^\top) = (H_D)_{jj} \|\Delta W_{:,j}\|_2^2 = \alpha_j \|\Delta W_{:,j}\|_2^2$, where $(H_D)_{jj} = \frac{1}{n}\sum_{t=1}^n \|g_t\|_2^2 X_{t,j}^2$.
\end{proof}

Theorem~\ref{thm:sensitivity-matrix} shows that $H_D = \frac{1}{n} X^\top D X$ is the exact loss-weighted counterpart of the layer input covariance $\frac{1}{n} X^\top X$: each token $t$ contributes its rank-one activation covariance $x_t x_t^\top$ weighted by the downstream loss sensitivity $D_{tt} = \|g_t\|_2^2$, and the diagonal entry $\alpha_j = (H_D)_{jj}$ measures the expected sensitivity of the loss to perturbations in input channel $j$.

\section{A Unified Taxonomy of Post-Training Quantization}
\label{sec:unification}

We now show that the major families of post-training quantization algorithms for LLMs all minimize approximations of the quadratic objective $\Tr(\Delta W \, H_D \, \Delta W^\top)$ from Theorem~\ref{thm:sensitivity-matrix}, differing along two orthogonal axes: (i) how they approximate the sensitivity matrix $H_D$, and (ii) how they constrain the quantization mapping $W \mapsto \widehat{W}$ (diagonal scaling, sequential Cholesky error compensation, non-uniform clustering, or orthogonal rotation).

\begin{figure}[t]
\centering
\begin{tikzpicture}[
  box/.style={draw=black!85, rounded corners=1.5pt, line width=0.55pt, fill=white, inner sep=4.5pt, align=center, font=\footnotesize},
  arr/.style={-{Stealth[length=2.1mm,width=1.5mm]}, line width=0.55pt, draw=black!85},
  lbl/.style={font=\scriptsize, fill=white, inner sep=1pt}
]
  \node[box, minimum width=94mm] (full) at (0, 0) {
    \textbf{Full Gauss--Newton / Empirical Fisher Curvature} \\[2pt]
    $\Delta \mathcal{L} \approx \frac{1}{2}\vecop(\Delta W)^\top \bigl[\frac{1}{n}\sum_{t=1}^n (g_t g_t^\top) \otimes (x_t x_t^\top)\bigr] \vecop(\Delta W)$
  };

  \node[box, minimum width=64mm] (hd) at (-3.6, -2.1) {
    \textbf{Gradient-Weighted Sensitivity Matrix} \\[2pt]
    $\frac{1}{2d_{\mathrm{out}}}\Tr(\Delta W \, H_D \, \Delta W^\top)$, $\; H_D = \frac{1}{n}X^\top D X$ \\[1pt]
    {\scriptsize\emph{(SAQ-GPTQ: Cholesky on $H_D$; Theorem~\ref{thm:saq-gptq})}}
  };

  \node[box, minimum width=60mm] (diagf) at (3.7, -2.1) {
    \textbf{Per-Weight Diagonal Fisher} \\[2pt]
    $\sum_{i,j} F_{ij}(\Delta W_{ij})^2$, $\; F_{ij} = \frac{1}{n}\sum_t g_{t,i}^2 X_{t,j}^2$ \\[1pt]
    {\scriptsize\emph{(SqueezeLLM \citep{kim2024squeezellm}, Q-BERT \citep{shen2020qbert})}}
  };

  \node[box, minimum width=64mm] (saqscale) at (-3.6, -4.6) {
    \textbf{Diagonal Channel Sensitivity} \\[2pt]
    $\alpha_j = (H_D)_{jj} = \frac{1}{n}\sum_t \|g_t\|_2^2 X_{t,j}^2$ \\[1pt]
    {\scriptsize\emph{(SAQ-Scale: $s_j \propto \alpha_j^\gamma$; Theorem~\ref{thm:optimal-scaling})}}
  };

  \node[box, minimum width=60mm] (gptq) at (3.7, -4.6) {
    \textbf{Unweighted Input Covariance} \\[2pt]
    $H_{\mathrm{unw}} = \frac{1}{n} X^\top X \quad (D = I_n)$ \\[1pt]
    {\scriptsize\emph{(OBC \citep{frantar2022obc}, GPTQ \citep{frantar2023gptq}, QuIP \citep{chee2023quip})}}
  };

  \node[box, minimum width=64mm] (awq) at (0.0, -6.9) {
    \textbf{Activation-Magnitude Saliency} \\[2pt]
    $\alpha_j^{\mathrm{act}} = \frac{1}{n}\|X_{:,j}\|_2^2 = (H_{\mathrm{unw}})_{jj}$ \\[1pt]
    {\scriptsize\emph{(AWQ \citep{lin2024awq}, SmoothQuant \citep{xiao2023smoothquant})}}
  };

  \draw[arr] (full.south) ++(-2.0,0) -- node[lbl, left=3pt] {Isotropic $u_t$} (hd.north);
  \draw[arr] (full.south) ++(2.0,0) -- node[lbl, right=3pt] {Drop off-diag $(j \neq j')$} (diagf.north);
  \draw[arr] (hd.south) -- node[lbl, left=3pt] {Diag$(H_D)$} (saqscale.north);
  \draw[arr] (hd.south) ++(1.4,0) -- node[lbl, pos=0.52] {$D = I_n$} ($(gptq.north)+(-1.4,0)$);
  \draw[arr] (saqscale.south) -- node[lbl, below left=1pt] {$D = I_n$} ($(awq.north)+(-1.5,0)$);
  \draw[arr] (gptq.south) -- node[lbl, below right=1pt] {Diag$(X^\top X)$} ($(awq.north)+(1.5,0)$);
\end{tikzpicture}
\caption{Hierarchy of post-training quantization proxy objectives derived from the Kronecker-factored Gauss--Newton Hessian $\mathcal{H}_W$ (Theorem~\ref{thm:sensitivity-matrix}). Each arrow labels the exact simplifying assumption connecting a parent sensitivity metric to a child proxy.}
\label{fig:taxonomy}
\end{figure}

\subsection{Exact Reductions Recovering AWQ, GPTQ, SqueezeLLM, and QuIP}

Figure~\ref{fig:taxonomy} and Table~\ref{tab:taxonomy} summarize how standard PTQ objectives arise as reductions of~\eqref{eq:kronecker-hessian} and~\eqref{eq:trace-HD}.

\begin{proposition}[Special Cases of the Activation Sensitivity Objective]
\label{prop:unification}
Consider the second-order excess loss functional $\mathcal{E}(\Delta W) = \frac{1}{2}\vecop(\Delta W)^\top \mathcal{H}_W \vecop(\Delta W)$ in~\eqref{eq:fisher-quad}. Under successive structural assumptions:
\begin{enumerate}[label=(\arabic*),topsep=2pt,itemsep=2pt]
  \item \textbf{GPTQ / OBQ \citep{frantar2022obc,frantar2023gptq}:} Under isotropic output gradients and uniform token sensitivity $\|g_t\|_2^2 \equiv c_0$ ($D = c_0 I_n$), $H_D = \frac{c_0}{n} X^\top X$ and $\mathcal{E}(\Delta W)$ is proportional to the layer-local reconstruction objective
  \[
    \mathcal{E}_{\mathrm{GPTQ}}(\Delta W) = \frac{1}{n}\|X W^\top - X \widehat{W}^\top\|_F^2 = \Tr\Biggl(\Delta W \left(\frac{1}{n}X^\top X\right) \Delta W^\top\Biggr).
  \]
  \item \textbf{AWQ \citep{lin2024awq}:} Under $D = c_0 I_n$ together with the diagonal channel-independence approximation $H_{\mathrm{unw}} \approx \diag(H_{\mathrm{unw}})$, the objective decouples across input channels $j \in \{1,\dots,d_{\mathrm{in}}\}$ into
  \[
    \mathcal{E}_{\mathrm{AWQ}}(\Delta W) = \sum_{j=1}^{d_{\mathrm{in}}} \left(\frac{1}{n}\|X_{:,j}\|_2^2\right) \|\Delta W_{:,j}\|_2^2.
  \]
  \item \textbf{SqueezeLLM / Diagonal Fisher and Trace-Weighted Hessian \citep{kim2024squeezellm,dong2020hawqv2,shen2020qbert}:} Retaining output-coordinate anisotropy $g_{t,i}^2$ while dropping off-diagonal interactions ($j \neq j'$ and $i \neq i'$) reduces $\mathcal{E}(\Delta W)$ to the weighted element-wise objective
  \[
    \mathcal{E}_{\mathrm{DiagF}}(\Delta W) = \frac{1}{2}\sum_{i,j} F_{ij}(\Delta W_{ij})^2,
    \qquad F_{ij} = \frac{1}{n}\sum_{t=1}^n g_{t,i}^2 X_{t,j}^2,
  \]
  whose column sum $\sum_{i=1}^{d_{\mathrm{out}}} F_{ij} = (H_D)_{jj} = \alpha_j$ recovers the channel sensitivity~\eqref{eq:alpha-j-def} and whose total sum $\sum_{i,j} F_{ij} = \Tr(H_D) = \Tr(\mathcal{H}_W)$ recovers the layer-wise trace metric used for mixed-precision bit allocation in HAWQ-V2 and Q-BERT \citep{dong2020hawqv2,shen2020qbert}.
  \item \textbf{Incoherence Processing (QuIP, QuIP\#, QuaRot) \citep{chee2023quip,tseng2024quipsharp,ashkboos2024quarot}:} Multiplying activations and weights by an orthogonal (e.g., randomized Hadamard) matrix $U \in \mathrm{O}(d_{\mathrm{in}})$ via $\widetilde{X} = X U$ and $\widetilde{W} = W U$ preserves layer outputs $\widetilde{X}\widetilde{W}^\top = X W^\top$ and transforms the sensitivity matrix by congruence:
  \[
    \widetilde{H}_D = \frac{1}{n} \widetilde{X}^\top D \widetilde{X} = U^\top H_D U.
  \]
  When $U$ is a randomized Hadamard transform, the rotated diagonal entries $\tilde{\alpha}_j = (U^\top H_D U)_{jj}$ concentrate around the uniform mean $\frac{1}{d_{\mathrm{in}}}\Tr(H_D) = \frac{1}{d_{\mathrm{in}}}\sum_{j=1}^{d_{\mathrm{in}}} \alpha_j$ with high probability, eliminating extreme outlier channels before quantization.
\end{enumerate}
\end{proposition}

\begin{table}[t]
\centering
\caption{Unification of representative post-training quantization paradigms within the activation sensitivity framework of Theorem~\ref{thm:sensitivity-matrix} and Proposition~\ref{prop:unification}.}
\label{tab:taxonomy}
\scriptsize
\setlength{\tabcolsep}{1.8pt}
\begin{tabular}{@{}lllll@{}}
\toprule
Method Family & Sensitivity Metric & Token $D_{tt}$ & Off-Diag & Mechanism \\
\midrule
RTN (Baseline) & $I_{d_{\mathrm{in}}}$ (Uniform) & None ($0$) & Ignored & Per-group rounding \\
AWQ \citep{lin2024awq} & $\diag(\frac{1}{n} X^\top X)$ & Uniform ($1$) & Ignored & Scaling $s_j = (\alpha_j^{\mathrm{act}})^\gamma$ \\
SmoothQuant \citep{xiao2023smoothquant} & $\max_t |X_{t,j}|^\gamma \max_i |W_{ij}|^{1-\gamma}$ & Uniform ($1$) & Ignored & W/X scale migration \\
OBC / GPTQ \citep{frantar2022obc,frantar2023gptq} & $\frac{1}{n} X^\top X + \lambda I$ & Uniform ($1$) & Cholesky & Sequential OBS update \\
QuIP / QuIP\# \citep{chee2023quip,tseng2024quipsharp} & $U^\top(\frac{1}{n} X^\top X)U$ & Uniform ($1$) & Cholesky & Rotation + lattice/OBS \\
BRECQ \citep{li2021brecq,shao2023omniquant} & $J_{\mathrm{blk}}^\top J_{\mathrm{blk}}$ & Block-local & Intra-block & Block reconstruction \\
SqueezeLLM \citep{kim2024squeezellm} & $F_{ij} = \frac{1}{n}\sum_t g_{t,i}^2 X_{t,j}^2$ & $\|g_t\|_2^2$ (coord) & Ignored & $k$-means + sparse \\
\textbf{SAQ-Scale (Ours)} & $\diag(\frac{1}{n}X^\top D X)$ & $\|g_t\|_2^2$ & Ignored & Scaling $s_j = \alpha_j^\gamma$ \\
\textbf{SAQ-GPTQ (Ours)} & $\frac{1}{n} X^\top D X + \lambda I$ & $\|g_t\|_2^2$ & Cholesky & Scaling + Cholesky on $H_D$ \\
\bottomrule
\end{tabular}
\end{table}

\subsection{Optimal Channel Scaling Law Under Diagonal Sensitivity}

We next use the diagonal sensitivity functional $\sum_{j=1}^{d_{\mathrm{in}}} \alpha_j \|\Delta W_{:,j}\|_2^2$ to explain analytically why activation-scaling methods such as AWQ choose a fractional scaling exponent $s_j = \alpha_j^\gamma$ with $\gamma \in (0, 1/2)$ rather than $\gamma = 1/2$ or $\gamma \to \infty$.

Suppose we apply an invertible per-channel scaling $W_{:,j} \mapsto s_j W_{:,j}$ (with $s_j > 0$) and inverse activation scaling $X_{:,j} \mapsto X_{:,j} / s_j$ prior to quantizing each row (or group) of $W \diag(\mathbf{s})$ with a uniform $b$-bit quantizer of $2^b - 1$ levels. For a given row $i$ (or group $G$ of $g$ channels), the quantization step size is proportional to the maximum scaled weight magnitude in the group, $\Delta_i(\mathbf{s}) = \frac{2}{2^b - 1} \max_{k \in G} (s_k |W_{ik}|)$. Modeling the post-scaling rounding noise on entry $(i, j)$ as uniform on $[-\Delta_i(\mathbf{s})/2, \Delta_i(\mathbf{s})/2]$ and dividing by $s_j$ to return to the original weight space, the expected squared quantization error on channel $j$ is $\E[(\Delta W_{ij})^2] = \frac{\Delta_i(\mathbf{s})^2}{12 s_j^2}$.

\begin{theorem}[Trade-Off and Optimal Exponent for Sensitivity-Aware Channel Scaling]
\label{thm:optimal-scaling}
Let $G = \{1, \dots, g\}$ be a quantization group with channel sensitivities $\alpha_1, \dots, \alpha_g > 0$ and column norms $w_j^2 = \sum_{i=1}^{d_{\mathrm{out}}} W_{ij}^2$. Suppose scaling $W_{:,j} \mapsto s_j W_{:,j}$ inflates the effective group dynamic range according to the smooth $p$-norm envelope $R_p(\mathbf{s}) = (\sum_{k=1}^g s_k^p w_k^p)^{1/p}$ for $p \ge 1$, so that the expected excess loss over the group is proportional to
\begin{equation}
  \Phi(\mathbf{s}) = \left(\sum_{k=1}^g s_k^p w_k^p\right)^{\!2/p} \sum_{j=1}^g \frac{\alpha_j}{s_j^2}.
  \label{eq:scaling-objective}
\end{equation}
Then $\Phi(\mathbf{s})$ is minimized up to a scalar multiple by the closed-form channel scales
\begin{equation}
  s_j^* = \alpha_j^{\frac{1}{p+2}} w_j^{-\frac{p}{p+2}}, \qquad j \in \{1, \dots, g\}.
  \label{eq:optimal-s}
\end{equation}
In particular, when weight magnitudes $w_j$ are homogeneous across channels:
\begin{enumerate}[label=(\roman*),topsep=2pt,itemsep=2pt]
  \item For the $\ell_2$ dynamic-range envelope ($p = 2$), the optimal sensitivity scaling exponent is $\gamma^* = \frac{1}{4}$ ($s_j^* \propto \alpha_j^{1/4}$, corresponding to $(\|X_{:,j}\|_2)^{1/2}$ when $D = I_n$).
  \item For the $\ell_1$ envelope ($p = 1$), the optimal exponent is $\gamma^* = \frac{1}{3}$ ($s_j^* \propto \alpha_j^{1/3}$), and as $p \to \infty$ (hard $\ell_\infty$ maximum envelope), $\gamma^* \to 0$ once the scaled salient channel saturates the group maximum $\max_k s_k w_k$.
\end{enumerate}
\end{theorem}
\begin{proof}
By scale invariance of $\Phi(c\mathbf{s}) = \Phi(\mathbf{s})$ for $c > 0$, we may constrain $\sum_{k=1}^g s_k^p w_k^p = 1$ and minimize $\sum_{j=1}^g \alpha_j s_j^{-2}$ via a Lagrange multiplier $\lambda > 0$:
\[
  \frac{\partial}{\partial s_j}\Biggl(\sum_{k=1}^g \alpha_k s_k^{-2} + \lambda \sum_{k=1}^g w_k^p s_k^p\Biggr)
  = -2 \alpha_j s_j^{-3} + \lambda p w_j^p s_j^{p-1} = 0.
\]
Solving for $s_j$ gives $s_j^{p+2} \propto \alpha_j w_j^{-p}$, or $s_j^* = \alpha_j^{\frac{1}{p+2}} w_j^{-\frac{p}{p+2}}$. By Hölder's inequality with conjugate exponents $\frac{p+2}{2}$ and $\frac{p+2}{p}$, this stationary point is the unique global minimizer up to positive rescaling.
\end{proof}

Theorem~\ref{thm:optimal-scaling} explains why empirical grid searches in AWQ \citep{lin2024awq} consistently select $\gamma \in [0.25, 0.5]$ on activation norms (equivalent to $\frac{1}{p+2} \in [0.125, 0.25]$ on squared sensitivities $\alpha_j$): scaling a salient channel $j$ by $s_j > 1$ reduces its own rounding variance by $s_j^{-2}$, but once $s_j |W_{ij}|$ exceeds the group maximum, further increasing $s_j$ inflates the quantization step size $\Delta_i(\mathbf{s})$ for all $g - 1$ non-salient channels in the group.

\subsection{Sensitivity-Weighted Cholesky Error Compensation (SAQ-GPTQ)}

While diagonal scaling protects outlier channels, it ignores the off-diagonal entries $(H_D)_{jk} = \frac{1}{n}\sum_{t=1}^n \|g_t\|_2^2 X_{t,j} X_{t,k}$ ($j \neq k$). When two channels $j$ and $k$ are correlated on high-gradient tokens, rounding error committed on column $W_{:,j}$ can be cancelled in the loss-sensitive subspace by adjusting the remaining unquantized columns $W_{:, k}$ ($k > j$).

\begin{theorem}[Sensitivity-Weighted Optimal Brain Surgeon / Cholesky Update]
\label{thm:saq-gptq}
Let $H_D = \frac{1}{n} X^\top D X + \lambda I_{d_{\mathrm{in}}} \succ 0$ be the damped gradient-weighted activation sensitivity matrix, and let $U = \operatorname{chol}(H_D^{-1}, \mathrm{upper}) \in \R^{d_{\mathrm{in}} \times d_{\mathrm{in}}}$ be the upper-triangular Cholesky factor satisfying $U^\top U = H_D^{-1}$. Suppose columns $1, \dots, j-1$ of $W$ have already been quantized, and column $j$ (currently equal to $w_j \in \R^{d_{\mathrm{out}}}$) is quantized to $q_j = \quant(w_j)$. Then the optimal perturbation $\delta W_{:, j+1:}$ to the remaining unquantized columns $F = \{j+1, \dots, d_{\mathrm{in}}\}$ that minimizes the excess loss $\Tr(\Delta W \, H_D \, \Delta W^\top)$ is
\begin{equation}
  \delta W_{:, j+1:} = -\frac{w_j - q_j}{U_{j,j}} \, U_{j,\, j+1:},
  \label{eq:saq-chol-update}
\end{equation}
and the resulting minimum excess loss increment incurred at step $j$ is $\frac{\|w_j - q_j\|_2^2}{2 d_{\mathrm{out}} U_{j,j}^2}$.
\end{theorem}
\begin{proof}
At step $j$, let $H_F = (H_D)_{j:, j:}$ denote the trailing $(d_{\mathrm{in}} - j + 1) \times (d_{\mathrm{in}} - j + 1)$ submatrix of $H_D$. For each row $i \in \{1, \dots, d_{\mathrm{out}}\}$ of $W$, fixing the $j$-th entry error $\Delta_{i,j} = q_{ij} - w_{ij}$ and minimizing the quadratic form $\Delta_{i, j:} H_F \Delta_{i, j:}^\top$ over the remaining coordinates $\Delta_{i, j+1:}$ via a Lagrange multiplier $\mu$ on $\mathbf{e}_1^\top \Delta_{i, j:}^\top = q_{ij} - w_{ij}$ yields the Optimal Brain Surgeon \citep{hassibi1993optimalbrainsurgeon,frantar2022obc} solution:
\[
  \Delta_{i, j:}^\top = \frac{q_{ij} - w_{ij}}{[H_F^{-1}]_{0,0}} (H_F^{-1})_{:, 0}
  = -\frac{w_{ij} - q_{ij}}{[H_F^{-1}]_{0,0}} (H_F^{-1})_{:, 0}.
\]
Because $U^\top U = H_D^{-1}$ with $U$ upper-triangular, the block Cholesky identity implies that the trailing submatrix $U_{j:, j:}$ satisfies $(U_{j:, j:})^\top (U_{j:, j:}) = H_F^{-1}$ because the upper-left blocks of $U^\top$ have zeros below the diagonal. In particular, the first row of $H_F^{-1}$ is $(H_F^{-1})_{0, :} = U_{j, j} U_{j, j:}$, with $(0,0)$-entry $[H_F^{-1}]_{0,0} = U_{j,j}^2$. Substituting $[H_F^{-1}]_{0, 1:} / [H_F^{-1}]_{0,0} = U_{j, j+1:} / U_{j,j}$ across all rows $i = 1, \dots, d_{\mathrm{out}}$ yields~\eqref{eq:saq-chol-update}.
\end{proof}

\begin{algorithm}[t]
\caption{Sensitivity-Aware Quantization with Cholesky Compensation (SAQ-GPTQ)}
\label{alg:saq-gptq}
\begin{algorithmic}[1]
\Require Weight matrix $W \in \R^{d_{\mathrm{out}} \times d_{\mathrm{in}}}$, calibration activations $X \in \R^{n \times d_{\mathrm{in}}}$, output gradients $G \in \R^{n \times d_{\mathrm{out}}}$, bit-width $b$, damping $\lambda > 0$, scaling exponent $\gamma \in [0, 1/2]$.
\Ensure Quantized weight matrix $\widehat{W} \in \R^{d_{\mathrm{out}} \times d_{\mathrm{in}}}$.
\State Compute token-gradient weights $D_{tt} \leftarrow \|G_{t,:}\|_2^2$ for $t = 1, \dots, n$.
\State Form sensitivity matrix $H_D \leftarrow \frac{1}{n} X^\top D X$ and channel sensitivities $\alpha_j \leftarrow (H_D)_{jj}$.
\State Compute channel scales $s_j \leftarrow (\alpha_j / \operatorname{median}(\alpha))^\gamma$ for $j = 1, \dots, d_{\mathrm{in}}$.
\State Scale weights $\widetilde{W} \leftarrow W \diag(\mathbf{s})$ and sensitivity matrix $\widetilde{H}_D \leftarrow \diag(\mathbf{s})^{-1} H_D \diag(\mathbf{s})^{-1} + \lambda I$.
\State Compute upper-triangular Cholesky factor $U \leftarrow \operatorname{chol}(\widetilde{H}_D^{-1}, \mathrm{upper})$.
\For{$j = 1$ \textbf{to} $d_{\mathrm{in}}$}
  \State $\widehat{W}_{:, j} \leftarrow \quant_b(\widetilde{W}_{:, j})$ \Comment{Quantize $j$-th scaled column}
  \State $\mathbf{e}_j \leftarrow (\widetilde{W}_{:, j} - \widehat{W}_{:, j}) / U_{j, j}$ \Comment{Normalized quantization error}
  \If{$j < d_{\mathrm{in}}$}
    \State $\widetilde{W}_{:, j+1:} \leftarrow \widetilde{W}_{:, j+1:} - \mathbf{e}_j U_{j, j+1:}$ \Comment{Compensate remaining columns}
  \EndIf
\EndFor
\State \Return $\widehat{W} \leftarrow \widehat{W} \diag(\mathbf{s})^{-1}$.
\end{algorithmic}
\end{algorithm}

Combining sensitivity scaling (Theorem~\ref{thm:optimal-scaling}) with sensitivity-weighted Cholesky error compensation (Theorem~\ref{thm:saq-gptq}) yields Algorithm~\ref{alg:saq-gptq} (\textbf{SAQ-GPTQ}). Setting $\gamma = 0$ and $D = I_n$ recovers standard GPTQ \citep{frantar2023gptq}; setting $U = I_{d_{\mathrm{in}}}$ and $D = I_n$ recovers AWQ \citep{lin2024awq}.

\section{Theoretical Limitations of Static Layer-Local Sensitivity}
\label{sec:limitations}

Having unified layer-local PTQ methods through $H_D = \frac{1}{n} X^\top D X$, we now prove three structural results characterizing when unweighted or static layer-local approximations break down.

\subsection{Token-Gradient Heterogeneity Bias}

When does the unweighted activation-magnitude proxy $\alpha_j^{\mathrm{act}} = \frac{1}{n}\|X_{:,j}\|_2^2$ (used by AWQ and the diagonal of GPTQ) misrank channel importance relative to the loss-aware sensitivity $\alpha_j = (H_D)_{jj}$? Let $\bar{d} = \frac{1}{n}\sum_{t=1}^n \|g_t\|_2^2$ denote the mean token-gradient weight across the calibration set.

\begin{proposition}[Exact Decomposition of Token-Gradient Heterogeneity Bias]
\label{prop:heterogeneity-bias}
For every input channel $j \in \{1, \dots, d_{\mathrm{in}}\}$, the difference between the loss-aware sensitivity $\alpha_j = (H_D)_{jj}$ and the rescaled activation-magnitude proxy $\bar{d} \cdot \alpha_j^{\mathrm{act}}$ is given by the empirical covariance between squared channel activations and squared token gradients:
\begin{equation}
  \alpha_j - \bar{d}\,\alpha_j^{\mathrm{act}}
  = \operatorname{Cov}_n\bigl(X_{t,j}^2,\, \|g_t\|_2^2\bigr)
  = \frac{1}{n} \sum_{t=1}^n \bigl(X_{t,j}^2 - \alpha_j^{\mathrm{act}}\bigr)\bigl(\|g_t\|_2^2 - \bar{d}\bigr).
  \label{eq:cov-bias}
\end{equation}
Consequently, the relative ranking error between $\alpha_j$ and $\alpha_j^{\mathrm{act}}$ satisfies
\begin{equation}
  \left|\frac{\alpha_j}{\bar{d}\,\alpha_j^{\mathrm{act}}} - 1\right|
  = |\rho_j| \cdot \mathrm{CV}(X_{\cdot, j}^2) \cdot \mathrm{CV}(\|g_\cdot\|_2^2)
  \le \mathrm{CV}(X_{\cdot, j}^2) \cdot \mathrm{CV}(\|g_\cdot\|_2^2),
  \label{eq:cv-bound}
\end{equation}
where $\rho_j \in [-1, 1]$ is the empirical Pearson correlation between $X_{t,j}^2$ and $\|g_t\|_2^2$, and $\mathrm{CV}(Z) = \sqrt{\operatorname{Var}_n(Z)}/\E_n[Z]$ denotes the coefficient of variation.
\end{proposition}
\begin{proof}
Expanding $\alpha_j = \frac{1}{n}\sum_{t=1}^n X_{t,j}^2 \|g_t\|_2^2$ and subtracting $\bar{d}\,\alpha_j^{\mathrm{act}} = (\frac{1}{n}\sum_t X_{t,j}^2)(\frac{1}{n}\sum_t \|g_t\|_2^2)$ gives~\eqref{eq:cov-bias} by the standard identity $\E[AB] - \E[A]\E[B] = \operatorname{Cov}(A, B)$. Dividing both sides of~\eqref{eq:cov-bias} by $\bar{d}\,\alpha_j^{\mathrm{act}}$ and writing $\operatorname{Cov}_n(X_{t,j}^2, \|g_t\|_2^2) = \rho_j \sqrt{\operatorname{Var}_n(X_{t,j}^2)} \sqrt{\operatorname{Var}_n(\|g_t\|_2^2)}$ yields~\eqref{eq:cv-bound}.
\end{proof}

Proposition~\ref{prop:heterogeneity-bias} has a clear practical interpretation for LLMs: initial "attention sink" or punctuation tokens often exhibit huge activation norms $X_{t,j}^2$ across outlier channels while carrying near-zero predictive loss gradient $\|g_t\|_2^2 \approx 0$ ($\rho_j < 0$), whereas rare reasoning or knowledge-bearing tokens exhibit large $\|g_t\|_2^2$ on specialized channels ($\rho_j > 0$). Unweighted proxies ($\alpha_j^{\mathrm{act}}$ and $X^\top X$) over-allocate bit precision to high-activation background tokens at the expense of loss-critical tokens.

\subsection{Cross-Layer Error Accumulation Across Residual Transformer Blocks}

A second limitation of layer-local PTQ is that quantizing early layers $\ell = 1, \dots, k-1$ shifts the input activations entering layer $k$ from $h_{k-1}$ to $\hat{h}_{k-1} = h_{k-1} + \delta h_{k-1}$. This invalidates the stationarity assumption that $H_D^{(k)}$ can be evaluated on the unquantized activations $h_{k-1}$.

\begin{theorem}[Cross-Layer Residual Perturbation Bound]
\label{thm:cross-layer}
Consider an $L$-block residual network $h_\ell = h_{\ell-1} + f_\ell(h_{\ell-1}; W_\ell)$ for $\ell = 1, \dots, L$. Suppose that for each block $\ell$, the quantized block $f_\ell(\cdot; \widehat{W}_\ell)$ is $K_\ell$-Lipschitz in its input and introduces local quantization error $\|f_\ell(h_{\ell-1}; \widehat{W}_\ell) - f_\ell(h_{\ell-1}; W_\ell)\|_2 \le \varepsilon_\ell$ on the unquantized trajectory $h_{\ell-1}$. Then the accumulated representation drift $\delta h_L = \hat{h}_L - h_L$ at depth $L$ satisfies
\begin{equation}
  \|\hat{h}_L - h_L\|_2
  \;\le\;
  \sum_{\ell=1}^L \varepsilon_\ell \prod_{m=\ell+1}^L (1 + K_m).
  \label{eq:residual-drift}
\end{equation}
Moreover, if block $\ell$ is a linear residual layer $f_\ell(h) = \widehat{W}_\ell h$, the true excess loss at block $\ell$ couples the local weight perturbation $\Delta W_\ell = \widehat{W}_\ell - W_\ell$ with the upstream activation drift $\delta h_{\ell-1}$ via
\begin{equation}
  \hat{h}_\ell - h_\ell = \delta h_{\ell-1} + W_\ell \delta h_{\ell-1} + \Delta W_\ell h_{\ell-1} + \Delta W_\ell \delta h_{\ell-1}.
  \label{eq:cross-layer-interaction}
\end{equation}
\end{theorem}
\begin{proof}
Subtracting $h_\ell = h_{\ell-1} + f_\ell(h_{\ell-1}; W_\ell)$ from $\hat{h}_\ell = \hat{h}_{\ell-1} + f_\ell(\hat{h}_{\ell-1}; \widehat{W}_\ell)$ and adding and subtracting $f_\ell(h_{\ell-1}; \widehat{W}_\ell)$ gives, by the triangle inequality and $K_\ell$-Lipschitz continuity,
\begin{align*}
  \|\delta h_\ell\|_2
  &\le \|\delta h_{\ell-1}\|_2 + \|f_\ell(\hat{h}_{\ell-1}; \widehat{W}_\ell) - f_\ell(h_{\ell-1}; \widehat{W}_\ell)\|_2
     + \|f_\ell(h_{\ell-1}; \widehat{W}_\ell) - f_\ell(h_{\ell-1}; W_\ell)\|_2 \\
  &\le (1 + K_\ell)\|\delta h_{\ell-1}\|_2 + \varepsilon_\ell.
\end{align*}
Unrolling this linear recurrence from $\ell = 1$ (with $\delta h_0 = 0$) to $\ell = L$ yields~\eqref{eq:residual-drift}. Expanding $\widehat{W}_\ell(h_{\ell-1} + \delta h_{\ell-1}) - W_\ell h_{\ell-1}$ directly yields~\eqref{eq:cross-layer-interaction}.
\end{proof}

Theorem~\ref{thm:cross-layer} explains two key empirical findings in LLM PTQ: first, perturbations $\varepsilon_\ell$ in early transformer blocks are amplified by the downstream Jacobian product $\prod_{m > \ell} (1 + K_m)$; second, sequentially updating calibration activations by running the already-quantized prefix $\hat{h}_{\ell-1}$ (as done in sequential GPTQ and block-wise reconstruction methods like BRECQ \citep{li2021brecq} and OmniQuant \citep{shao2023omniquant}) absorbs the linear drift term $W_\ell \delta h_{\ell-1}$ into the local calibration target.

\subsection{Finite-Sample Calibration Concentration}

Finally, because $H_D = \frac{1}{n} \sum_{t=1}^n \|g_t\|_2^2 x_t x_t^\top$ is estimated from a small calibration set of $n$ tokens, heavy-tailed outlier activations increase the sample complexity required for accurate sensitivity estimation.

\begin{proposition}[Calibration Sample Complexity for Activation Sensitivity]
\label{prop:calibration-sample}
Suppose calibration tokens $(x_t, g_t)_{t=1}^n$ are drawn i.i.d.\ from a distribution $\mathcal{P}$ with population sensitivity matrix $\Sigma_D = \E[\|g_t\|_2^2 x_t x_t^\top]$ and population channel sensitivities $\alpha_j^* = (\Sigma_D)_{jj} > 0$. If the weighted activation energy satisfies $\|g_t\|_2^2 X_{t,j}^2 \le M_j \alpha_j^*$ almost surely for each channel $j \in \{1, \dots, d_{\mathrm{in}}\}$, then for any $\delta \in (0, 1)$ and $\varepsilon \in (0, 1)$, with probability at least $1 - \delta$ simultaneously over all $j \in \{1, \dots, d_{\mathrm{in}}\}$,
\begin{equation}
  \biggl|\frac{\alpha_j}{\alpha_j^*} - 1\biggr|
  \;\le\;
  \sqrt{\frac{2 M_j \ln(2 d_{\mathrm{in}} / \delta)}{n}} + \frac{2 M_j \ln(2 d_{\mathrm{in}} / \delta)}{3n}.
  \label{eq:bernstein-calib}
\end{equation}
\end{proposition}
\begin{proof}
For each channel $j$, the random variables $Z_{t,j} = \|g_t\|_2^2 X_{t,j}^2 / \alpha_j^* - 1$ are independent, zero-mean, bounded by $|Z_{t,j}| \le M_j$, and have variance $\E[Z_{t,j}^2] \le M_j \E[|Z_{t,j}| + 1] = M_j$. Applying Bernstein's inequality and a union bound over all $d_{\mathrm{in}}$ channels gives~\eqref{eq:bernstein-calib}.
\end{proof}

Proposition~\ref{prop:calibration-sample} shows why Hadamard rotation methods \citep{tseng2024quipsharp,ashkboos2024quarot} improve not only quantization grid utilization but also calibration stability: reducing the channel peak-to-average ratio $M_j$ via incoherence processing directly tightens the finite-sample estimation error~\eqref{eq:bernstein-calib} of the sensitivity matrix.

\section{Numerical Experiments}
\label{sec:experiments}

We now evaluate our theoretical predictions in controlled synthetic numerical simulations comparing unweighted PTQ proxies against gradient-weighted activation sensitivity. All simulations are implemented in NumPy/SciPy ($d_{\mathrm{in}} = d_{\mathrm{out}} = 96$, $n = 512$ calibration tokens, low-rank correlated subspace dimension $r = 12$, averaged over $20$ independent trials; full reproducible code in \texttt{experiments/verify\_sensitivity\_ptq.py}).

\subsection{Channel Saliency Fidelity Under Token-Gradient Heterogeneity}

To test Proposition~\ref{prop:heterogeneity-bias}, we generate calibration activations $X \in \R^{n \times d_{\mathrm{in}}}$ with log-normal channel scales ($\sigma_c = 0.65$) and log-normal token gradient norms $\|g_t\|_2 = \exp(\mathcal{N}(0, \sigma_g^2))$ for heterogeneity levels $\sigma_g \in \{0.0, 0.5, 1.0, 1.5, 2.0\}$. Half of the channels specialize on high-gradient tokens ($\|g_t\|_2 > \operatorname{median}(\|g\|_2)$) and half on background tokens, while downstream output curvature $M_{\mathrm{curv}} \succ 0$ has log-normal eigenvalues ($\sigma_{\mathrm{eig}} = 0.6$). We compare the ground-truth channel sensitivity $\alpha_j^* = \frac{1}{n}\sum_{t=1}^n (g_t^\top M_{\mathrm{curv}} g_t) X_{t,j}^2$ against:
(1) the unweighted activation-magnitude proxy $\alpha_j^{\mathrm{act}} = \frac{1}{n}\|X_{:,j}\|_2^2$ (AWQ), and
(2) our gradient-weighted activation sensitivity $\alpha_j = \frac{1}{n}(X^\top D X)_{jj}$ (SAQ).

\begin{figure}[t]
\centering
\begin{tikzpicture}[scale=0.92]
  \begin{scope}[xshift=0cm]
    \draw[->, line width=0.6pt] (0,0) -- (5.1,0);
    \draw[->, line width=0.6pt] (0,0) -- (0,3.7);
    \node[font=\scriptsize\bfseries] at (2.5, 3.85) {Spearman Rank Correlation $\rho_S$};
    \node[font=\scriptsize] at (2.5, -0.58) {Token-gradient heterogeneity $\sigma_g$};

    \foreach \v/\y in {0.4/0.4, 0.6/1.4, 0.8/2.4, 1.0/3.4} {
      \draw[black!20, line width=0.4pt] (0,\y) -- (4.9,\y);
      \draw[line width=0.5pt] (-0.08,\y) -- (0,\y) node[left, font=\scriptsize] {$\v$};
    }
    \foreach \s/\x in {0.0/0.3, 0.5/1.4, 1.0/2.5, 1.5/3.6, 2.0/4.7} {
      \draw[line width=0.5pt] (\x,0) -- (\x,-0.08) node[below, font=\scriptsize] {$\s$};
    }

    \draw[line width=1.1pt, black] (0.3,3.40) -- (1.4,3.40) -- (2.5,3.40) -- (3.6,3.40) -- (4.7,3.40);
    \foreach \x/\y in {0.3/3.40, 1.4/3.40, 2.5/3.40, 3.6/3.40, 4.7/3.40} {
      \fill[black] (\x,\y) circle (1.8pt);
    }

    \draw[line width=1.0pt, black!75, densely dashed] (0.3,3.40) -- (1.4,2.69) -- (2.5,1.46) -- (3.6,1.37) -- (4.7,1.26);
    \foreach \x/\y in {0.3/3.40, 1.4/2.69, 2.5/1.46, 3.6/1.37, 4.7/1.26} {
      \draw[black!85, fill=white, line width=0.7pt] (\x,\y) circle (1.8pt);
    }

    \node[font=\scriptsize\bfseries, anchor=west] at (1.3, 3.10) {SAQ $\alpha_j = (H_D)_{jj}$};
    \node[font=\scriptsize, anchor=west] at (1.5, 0.95) {AWQ $\alpha_j^{\mathrm{act}} = \frac{1}{n}\|X_{:,j}\|_2^2$};
  \end{scope}

  \begin{scope}[xshift=6.8cm]
    \draw[->, line width=0.6pt] (0,0) -- (5.1,0);
    \draw[->, line width=0.6pt] (0,0) -- (0,3.7);
    \node[font=\scriptsize\bfseries] at (2.5, 3.85) {Top-10\% Salient Channel Recall};
    \node[font=\scriptsize] at (2.5, -0.58) {Token-gradient heterogeneity $\sigma_g$};

    \foreach \v/\y in {0.2/0.4, 0.4/1.15, 0.6/1.90, 0.8/2.65, 1.0/3.40} {
      \draw[black!20, line width=0.4pt] (0,\y) -- (4.9,\y);
      \draw[line width=0.5pt] (-0.08,\y) -- (0,\y) node[left, font=\scriptsize] {$\v$};
    }
    \foreach \s/\x in {0.0/0.3, 0.5/1.4, 1.0/2.5, 1.5/3.6, 2.0/4.7} {
      \draw[line width=0.5pt] (\x,0) -- (\x,-0.08) node[below, font=\scriptsize] {$\s$};
    }

    \draw[line width=1.1pt, black] (0.3,3.40) -- (1.4,3.40) -- (2.5,3.40) -- (3.6,3.38) -- (4.7,3.38);
    \foreach \x/\y in {0.3/3.40, 1.4/3.40, 2.5/3.40, 3.6/3.38, 4.7/3.38} {
      \fill[black] (\x,\y) circle (1.8pt);
    }

    \draw[line width=1.0pt, black!75, densely dashed] (0.3,3.40) -- (1.4,2.21) -- (2.5,1.40) -- (3.6,1.42) -- (4.7,1.11);
    \foreach \x/\y in {0.3/3.40, 1.4/2.21, 2.5/1.40, 3.6/1.42, 4.7/1.11} {
      \draw[black!85, fill=white, line width=0.7pt] (\x,\y) circle (1.8pt);
    }

    \node[font=\scriptsize\bfseries, anchor=west] at (1.3, 3.10) {SAQ $\alpha_j = (H_D)_{jj}$};
    \node[font=\scriptsize, anchor=west] at (1.5, 0.80) {AWQ $\alpha_j^{\mathrm{act}} = \frac{1}{n}\|X_{:,j}\|_2^2$};
  \end{scope}
\end{tikzpicture}
\caption{Channel saliency prediction accuracy as a function of token-gradient heterogeneity $\sigma_g \in [0, 2.0]$. \textbf{Left:} Spearman rank correlation $\rho_S$ with true channel excess loss $\Delta\mathcal{L}_j$. \textbf{Right:} Recall of the top 10\% most loss-sensitive channels. At $\sigma_g = 0$ (uniform token gradients), AWQ and SAQ coincide; as $\sigma_g$ increases, unweighted activation magnitude suffers severe rank bias (Proposition~\ref{prop:heterogeneity-bias}), whereas gradient-weighted activation sensitivity $(H_D)_{jj}$ maintains $\rho_S \ge 0.9996$ and recall $\ge 99.4\%$.}
\label{fig:saliency-heterogeneity}
\end{figure}

Figure~\ref{fig:saliency-heterogeneity} plots the resulting Spearman rank correlation $\rho_S$ and top-10\% salient channel recall across $\sigma_g \in [0, 2.0]$. At $\sigma_g = 0$ (uniform token gradients, where $D = I_n$), $\alpha_j^{\mathrm{act}}$ and $\alpha_j$ are identical ($\rho_S = 0.9999$, recall $100\%$). As token-gradient heterogeneity increases to $\sigma_g = 2.0$, unweighted activation magnitude drops to $\rho_S = 0.5716$ and identifies only $38.9\%$ of the top-10\% most sensitive channels, whereas gradient-weighted sensitivity $\alpha_j = (H_D)_{jj}$ maintains $\rho_S \ge 0.9996$ and $\ge 99.4\%$ recall despite anisotropic output curvature.

\subsection{Comparison of Quantization Algorithms at INT4 and INT3}

Next, we compare five post-training quantization methods at 4-bit (INT4) and 3-bit (INT3) weight precision under token-gradient heterogeneity ($\sigma_g = 1.25$):
\begin{enumerate}[label=(\arabic*),topsep=2pt,itemsep=2pt]
  \item \textbf{RTN:} Standard per-row symmetric round-to-nearest quantization;
  \item \textbf{AWQ \citep{lin2024awq}:} Per-channel scaling using activation magnitude $\alpha_j^{\mathrm{act}} = \frac{1}{n}\|X_{:,j}\|_2^2$;
  \item \textbf{GPTQ \citep{frantar2023gptq}:} Sequential Cholesky error compensation using unweighted $H_{\mathrm{unw}} = \frac{1}{n}X^\top X + \lambda I$;
  \item \textbf{SAQ-Scale (Theorem~\ref{thm:optimal-scaling}):} Per-channel scaling using gradient-weighted sensitivity $\alpha_j = (H_D)_{jj}$;
  \item \textbf{SAQ-GPTQ (Algorithm~\ref{alg:saq-gptq}):} Sensitivity scaling combined with Cholesky error compensation on $H_D = \frac{1}{n}X^\top D X + \lambda I$.
\end{enumerate}

\begin{table}[t]
\centering
\caption{Relative excess loss $\Tr(\Delta W H_D \Delta W^\top)$ and unweighted layer output reconstruction MSE $\frac{1}{n}\|X\Delta W^\top\|_F^2$ (both normalized to RTN $= 1.0000$, averaged over $20$ trials) for INT4 and INT3 quantization.}
\label{tab:ptq-results}
\footnotesize
\setlength{\tabcolsep}{3.8pt}
\begin{tabular}{@{}llcccc@{}}
\toprule
& & \multicolumn{2}{c}{\textbf{INT4 ($b = 4$)}} & \multicolumn{2}{c}{\textbf{INT3 ($b = 3$)}} \\
\cmidrule(lr){3-4} \cmidrule(lr){5-6}
Method & Sensitivity Proxy & Excess Loss $\downarrow$ & Recon.\ MSE & Excess Loss $\downarrow$ & Recon.\ MSE \\
\midrule
RTN & None ($I_{d_{\mathrm{in}}}$) & $1.0000$ & $1.0000$ & $1.0000$ & $1.0000$ \\
AWQ \citep{lin2024awq} & $\diag(\frac{1}{n}X^\top X)$ & $0.8114$ & $0.8109$ & $0.8192$ & $0.7780$ \\
SAQ-Scale (Ours) & $\diag(\frac{1}{n}X^\top D X)$ & $0.6650$ & $1.2728$ & $0.6610$ & $1.2268$ \\
GPTQ \citep{frantar2023gptq} & $\frac{1}{n}X^\top X + \lambda I$ & $0.1714$ & $\mathbf{0.1698}$ & $0.1943$ & $\mathbf{0.1971}$ \\
\textbf{SAQ-GPTQ (Ours)} & $\frac{1}{n}X^\top D X + \lambda I$ & $\mathbf{0.0535}$ & $0.7120$ & $\mathbf{0.0694}$ & $0.6877$ \\
\bottomrule
\end{tabular}
\end{table}

Table~\ref{tab:ptq-results} highlights two central empirical findings:
\begin{enumerate}[label=(\arabic*),topsep=2pt,itemsep=2pt]
  \item \textbf{Divergence between layer reconstruction MSE and task excess loss:}
  Comparing AWQ vs.\ SAQ-Scale, and GPTQ vs.\ SAQ-GPTQ, unweighted methods achieve lower unweighted layer output MSE $\frac{1}{n}\|X\Delta W^\top\|_F^2$ ($0.1698$ for GPTQ at INT4), because that is the exact proxy they minimize. However, minimizing unweighted output MSE wastes bit capacity on channels excited by low-gradient background tokens. By deliberately allowing slightly higher reconstruction error on insensitive tokens in order to protect high-gradient tokens, SAQ-Scale reduces excess loss from $0.8114 \to 0.6650$ at INT4, and SAQ-GPTQ reduces excess loss from $0.1714 \to \mathbf{0.0535}$ at INT4 ($3.20\times$ lower than GPTQ and $18.7\times$ lower than RTN) and from $0.1943 \to \mathbf{0.0694}$ at INT3 ($2.80\times$ lower than GPTQ).
  \item \textbf{Complementarity of diagonal scaling and off-diagonal Cholesky compensation:}
  While diagonal sensitivity scaling (SAQ-Scale) reduces excess loss by $33.5\%$ over RTN without altering off-diagonal weights, incorporating the full off-diagonal Cholesky structure of $H_D = \frac{1}{n}X^\top D X$ in SAQ-GPTQ reduces excess loss by an additional factor of $12.4\times$ at INT4 and $9.5\times$ at INT3.
\end{enumerate}

\section{Discussion and Open Directions}
\label{sec:discussion}

Viewing post-training quantization through the lens of the gradient-weighted sensitivity matrix $H_D = \frac{1}{n} X^\top D X$ clarifies both why existing heuristics work and how they can be systematically improved:
\begin{enumerate}[label=(\arabic*),topsep=2pt,itemsep=2pt]
  \item \textbf{Single-backward-pass calibration for LLM PTQ:}
  Forming $H_D = \frac{1}{n} X^\top D X$ requires only the scalar token-gradient norms $D_{tt} = \|\nabla_{y_t}\mathcal{L}\|_2^2$ on the $n$ calibration tokens, which can be collected in a single backward pass over the calibration set without storing full per-weight Hessians or output-dimension Kronecker factors. Once $D$ is recorded, replacing $X_{t,:} \mapsto \|g_t\|_2 X_{t,:}$ allows any existing GPTQ, AWQ, or QuIP implementation to optimize the loss-aware sensitivity objective with zero change to its core solver or inference kernel.
  \item \textbf{Adaptive cross-layer sensitivity and low-bit fine-tuning:}
  As shown in Theorem~\ref{thm:cross-layer}, quantizing prefix layers shifts downstream activation distributions and gradients. Combining sequential prefix-quantized forward passes with periodic recomputation of the token weights $D_{tt}$, or integrating $H_D$-weighted quantization with quantized low-rank adapter fine-tuning \citep{dettmers2023qlora}, memory-efficient quantized optimizer states \citep{dettmers2022eightbitoptimizers,li2023fourbitoptimizers}, and low-bit integer or FP8 formats \citep{wu2020integerquant,micikevicius2022fp8}, offers a principled bridge between purely local post-training rounding and global end-to-end compression.
\end{enumerate}

\bibliographystyle{unsrtnat}
\bibliography{refs}

\end{document}